\documentclass[letterpaper, 10 pt, conference]{ieeeconf}  

\IEEEoverridecommandlockouts                              

\ifdefined\pdfminorversion\pdfminorversion=7\fi 

\usepackage{graphicx}
\usepackage{graphics} 
\usepackage{epsfig} 
\usepackage{mathptmx} 
\usepackage{times} 
\usepackage{amsmath} 
\usepackage{amssymb}  
\usepackage{booktabs}
\usepackage{tabularx}
\usepackage{url}

\title{\LARGE \bf
Adaptive Question Generation for Ambiguous Robot Instructions \\Based on Information Gain and Human Cognitive Characteristics
}

\author{Anonymous Author(s)}

\author{Ryo Terashima$^{1}$, Yuga Yano$^{1}$ and Hakaru Tamukoh$^{1, 2}$
\thanks{This paper is based on results obtained from project JPNP16007 commissioned by the New Energy and Industrial Technology Development Organization (NEDO). This work was supported by JSPS KAKENHI Grant Numbers 23H03468 and 23K18495. This work was supported JST ALCA-Next Grant Number JPMJAN23F3. This work was supported by JST SPRING, Japan Grant Number JPMJSP2154.}
\thanks{All authors are with Kyushu Institute of Technology, Fukuoka, Japan.
        {\tt\small terashima.ryo631@mail.kyutech.jp, yano.yuuga158@mail.kyutech.jp} and {\tt\small tamukoh@brain.kyutech.ac.jp}
        }
\thanks{Hakaru Tamukoh is also affiliated with the Research Center for Neuromorphic AI Hardware, Fukuoka Japan.}%
}

\begin{document}

\maketitle

\thispagestyle{empty}
\pagestyle{empty}


\begin{abstract}
In this study, we propose an adaptive question generation mechanism that addresses uncertainty arising from ambiguous instructions.
Natural-language instructions for service robots often contain ambiguities.
In scenes with multiple similar objects, the robot must identify the target object by asking questions.
However, existing question generation methods focus on probabilistic discrimination efficiency and do not sufficiently account for user ease of response.
Asking questions without considering the user's ease of response may cause incorrect or uncertain answers. 
Such incorrect or uncertain answers lead to additional clarification turns and degraded dialogue efficiency.
The proposed mechanism integrates probabilistic discrimination efficiency and human cognitive characteristics to generate questions that users can intuitively and easily answer.
First, the proposed mechanism extracts visual attributes of candidate objects using a vision-language model (VLM).
Next, the proposed mechanism computes the information gain (IG) of each attribute and selects the visual attribute (e.g., color, shape) that most reduces uncertainty in the candidate set.
When multiple attributes have the same IG, the proposed mechanism applies a visual attribute filter based on cognitive characteristics to select the visual attribute that users can most intuitively perceive.
The proposed mechanism then prompts the VLM with the selected attribute and the corresponding attribute values of each candidate object to generate a concise natural-language question.
This process is repeated until the target object is uniquely identified.
The proposed mechanism achieved a target identification success rate of 88.5\% in the evaluation using a dataset of 120 scenes.
Furthermore, the proposed mechanism received high ratings in subjective evaluations of response ease.
\end{abstract}

\section{INTRODUCTION}

In recent years, the demand for service robots that coexist with humans in homes and commercial facilities has grown, driven by the declining birthrate, aging population, and shrinking workforce~\cite{Ono2022, YanoWaitersystem2023, holland2021service, acar2024restaurant, Isomoto_SII, terashima2026grid, YanoROMAN2024}.
Furthermore, robots require the ability to understand natural language instructions from users and handle various situations~\cite{Wang2024GPSR, memmesheimer2024EGPSR, SayCan}.
However, natural language instructions from users often involve ambiguity, as target objects are not always specified~\cite{misra2016tell, liang2024introspective, zhang2025clarify}.
For example, an instruction such as ``bring me a drink'' specifies only the target object's category.
Therefore, when multiple drinks exist in the environment, the robot cannot identify the target.
To select appropriate actions, the robot must obtain additional information from the user and resolve the ambiguity.

\begin{figure}[tb]
   \centering
   \includegraphics[width=\linewidth]{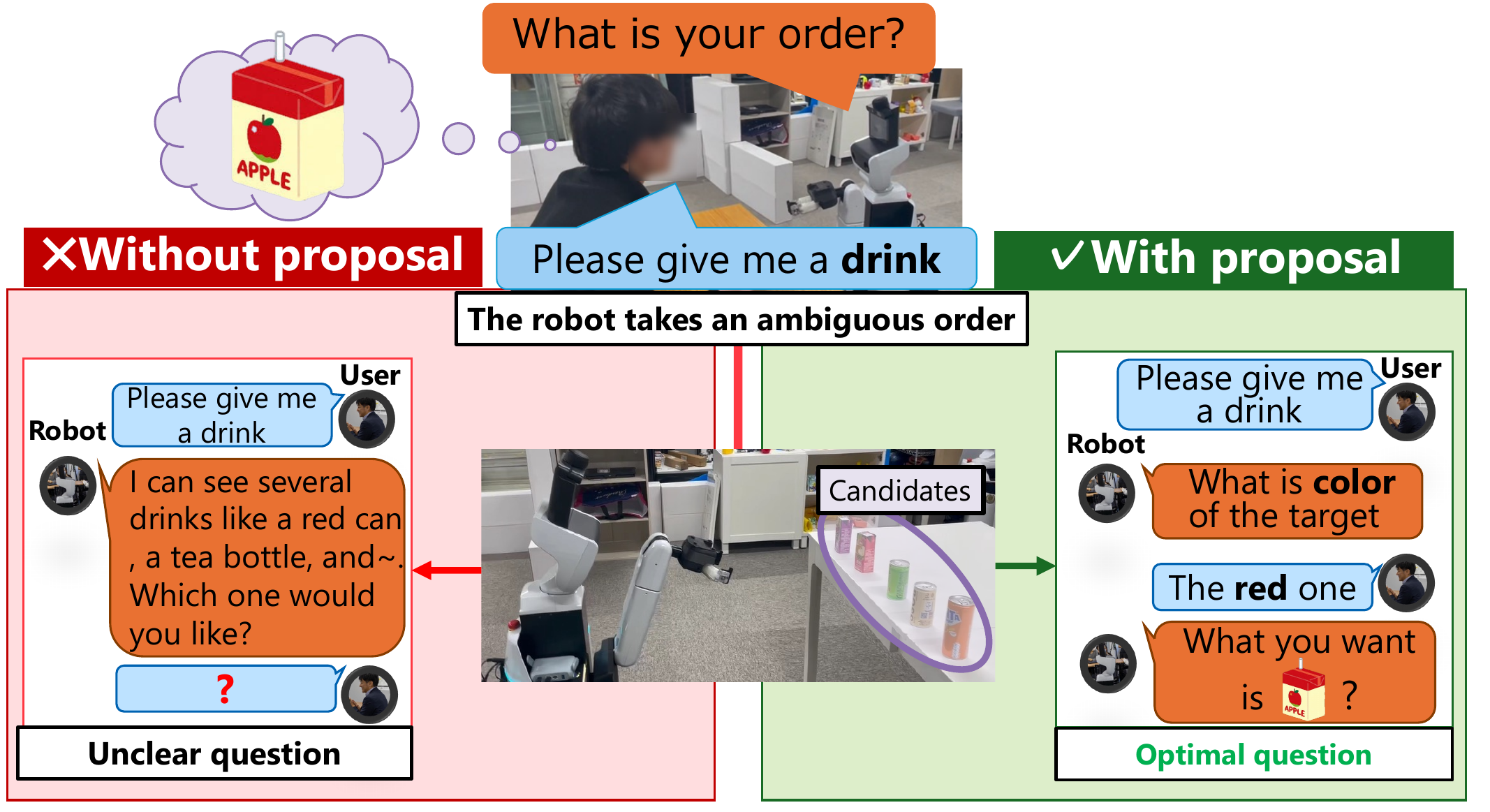}
  \caption{A scene in which a robot receives an ambiguous instruction and attempts to identify the target object from multiple candidate objects.
Left (red region, Without Proposed): The case without the proposed mechanism, where the user hesitates to answer and the target object cannot be identified.
Right (green region, With Proposed): The case with the proposed mechanism, where a concise question enables efficient ambiguity resolution.}
   \label{Intro_Fig}
\end{figure}


Ren et al. proposed KnowNo to address uncertainties arising from such ambiguous instructions and situations~\cite{KNOWNO}.
KnowNo quantifies the uncertainty before task execution and asks the user for clarification when ambiguity exists.
This approach focuses on determining when to ask questions to resolve ambiguity.
However, KnowNo does not address how to generate questions that users can easily answer.

Regarding question content, an approach that generates questions based on the probabilistic discrimination efficiency of candidate objects has been proposed~\cite{yang2022interactive}.
This study demonstrates the effectiveness of selecting visual attributes that efficiently partition the candidate object set.
However, attributes with high discrimination efficiency are not always easy for users to answer.
Even among attributes with the same discriminability, when users cannot intuitively recall those attributes, response errors and ambiguities may occur.
Response errors and ambiguities can lead to additional clarifications and increased user dialogue.
Human cognitive characteristics form the background of this issue.
When humans recognize and memorize objects, they do not recall all visual attributes equally.
There is a human tendency to recall visually salient features such as color and shape~\cite{treisman1980feature}.
Therefore, selecting attributes for questions requires consideration not only of probabilistic discrimination efficiency but also of user cognitive characteristics.
However, an approach that integrates probabilistic discrimination efficiency and human cognitive characteristics has not been explored.
Based on the above, this study establishes the following hypothesis.
\textbf{Hypothesis}: ``Integrating probabilistic discrimination efficiency and human cognitive characteristics enables the generation of questions that achieve high discrimination efficiency and are easy for users to answer.''

Based on this hypothesis, we propose an adaptive question generation mechanism that integrates information gain (IG) and cognitive characteristics.
Fig.~\ref{Intro_Fig} shows a comparative scenario in which a robot receives an ambiguous instruction.
Without the proposed mechanism, the robot generates a question that is unclear, making it difficult for the user to answer.
With the proposed mechanism, the robot generates a concise question focused on a single visual attribute with high discrimination efficiency, enabling the user to answer intuitively.

In this study, we conducted large-scale user experiments using crowdsourcing to quantitatively verify the discrimination efficiency and ease of response of the proposed mechanism.
In this study, we collect an original dataset of 120 scenes collected on a university campus.
The experimental results show that the proposed mechanism achieves a high target identification success rate and high ease of response for users.

The main contributions of this study are the following two points.

\begin{itemize}
\item Proposal of an adaptive question generation mechanism that integrates information gain and cognitive characteristics
\item Quantitative demonstration that human cognitive characteristics are effective for intuitive question generation.
\end{itemize}

\section{RELATED WORKS}
In this study, we address the problem of generating questions that are both efficient and easy for users to answer in response to ambiguous object search instructions.
To clarify the positioning of the proposed mechanism, we first review approaches to handling ambiguous instructions, then examine question-strategy design in dialogue.
Finally, discuss question design based on human cognitive characteristics.

\subsection{Handling ambiguous instructions}
\label{exsample_knowno}
Ren et al. proposed KnowNo as a method for handling ambiguous natural language instructions~\cite{KNOWNO}.
KnowNo takes user instructions and object recognition results from the environment as input, and generates multiple action candidates using a large language model (LLM)~\cite{anil2023palm2technicalreport}.
Then, KnowNo applies Conformal Prediction (CP) based on the likelihood of each candidate.
CP is a statistical method that constructs a prediction set that contains the correct answer with a given confidence level.
When the prediction set contains a single candidate, the robot proceeds with the execution action.
When multiple candidates remain, the robot asks the user for help to avoid taking an incorrect action.
In this way, KnowNo interacts with the user only when necessary and addresses ambiguous situations.
KnowNo achieves a high task success rate in both simulated and real environments.
However, KnowNo assumes that the user will always provide a correct answer to each question.
Therefore, KnowNo does not address how to generate questions that are easy for users to answer.

\subsection{Question strategy design in dialogue}
As question strategies for ambiguous instructions, methods that minimize the entropy of the candidate set and approaches that maximize IG have been proposed~\cite{yang2022interactive, deits2013clarifying}.
For example, Deits et al. demonstrated that selected questions with high discrimination efficiency based on entropy minimization can quantitatively reduce uncertainty with each dialogue turn~\cite{deits2013clarifying}.
Furthermore, Yang et al. proposed a method that progressively resolves ambiguity using visual attributes such as object color~\cite{yang2022interactive}.
Yang et al. achieved a target identification accuracy of 91\% in real-world experiments, confirming that attribute-based questions are effective for object identification.

These studies demonstrate the effectiveness of question selection based on probabilistic efficiency.
However, an attribute that efficiently partitions the candidate set is not necessarily easy for users to answer.
This gap between discrimination efficiency and user ease of response has not been sufficiently explored.
In this study, we investigate whether considering users' ease of response when selecting questions leads to more effective ambiguity resolution.

\begin{table}[tb]
\centering
\caption{Definition of visual attributes used in this study}
\label{tab:visual_attributes_definition}
\begin{tabular}{l p{6cm}}
\hline
Attribute & Definition \\
\hline
Color &  Represents the primary color of the object \\
Size &  Represents the relative size among candidate objects \\
Shape & Represents the overall form and outline of the object\\
Text &  Represents textual information written on the surface of the object \\
Material & Represents the material of the object estimated from its appearance \\
Illustration &  Represents the presence and content of patterns or illustrations drawn on the surface of the object \\
\hline
\end{tabular}
\end{table}

\subsection{Question design based on human cognitive characteristics}
When humans recognize and memorize objects, they do not memorize all visual attributes equally.
Visually salient features such as color and shape tend to be more memorable and preferentially recalled~\cite{treisman1980feature}.
Based on these findings, SalBench~\cite{salbench} defines visual attributes that humans can intuitively perceive.
These findings indicate that selecting attributes for questions requires not only discrimination efficiency but also a cognitive perspective on whether users can intuitively recall and answer them easily.

However, existing studies have not sufficiently explored an approach that defines attributes suitable for questions based on human cognitive characteristics and incorporates them into question design.
Focusing on this point, we propose a framework that integrates cognitive characteristics into question design.

\begin{figure*}[t]
   \centering
   \includegraphics[width=\linewidth]{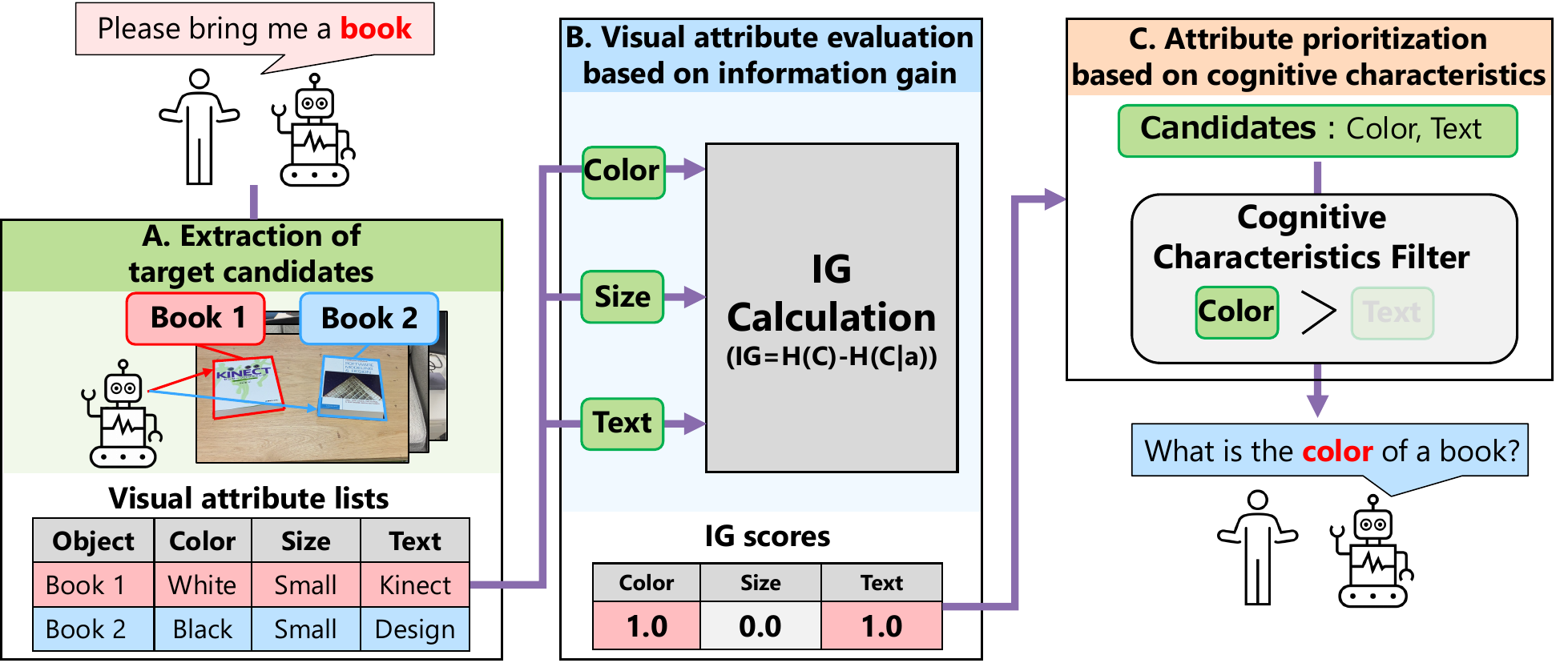} 
   \caption{Overview of the adaptive question generation mechanism based on information gain and human cognitive characteristics.}
   \label{proposed_question}
\end{figure*}

\begin{figure}[tb]
   \centering
   \includegraphics[width=\linewidth]{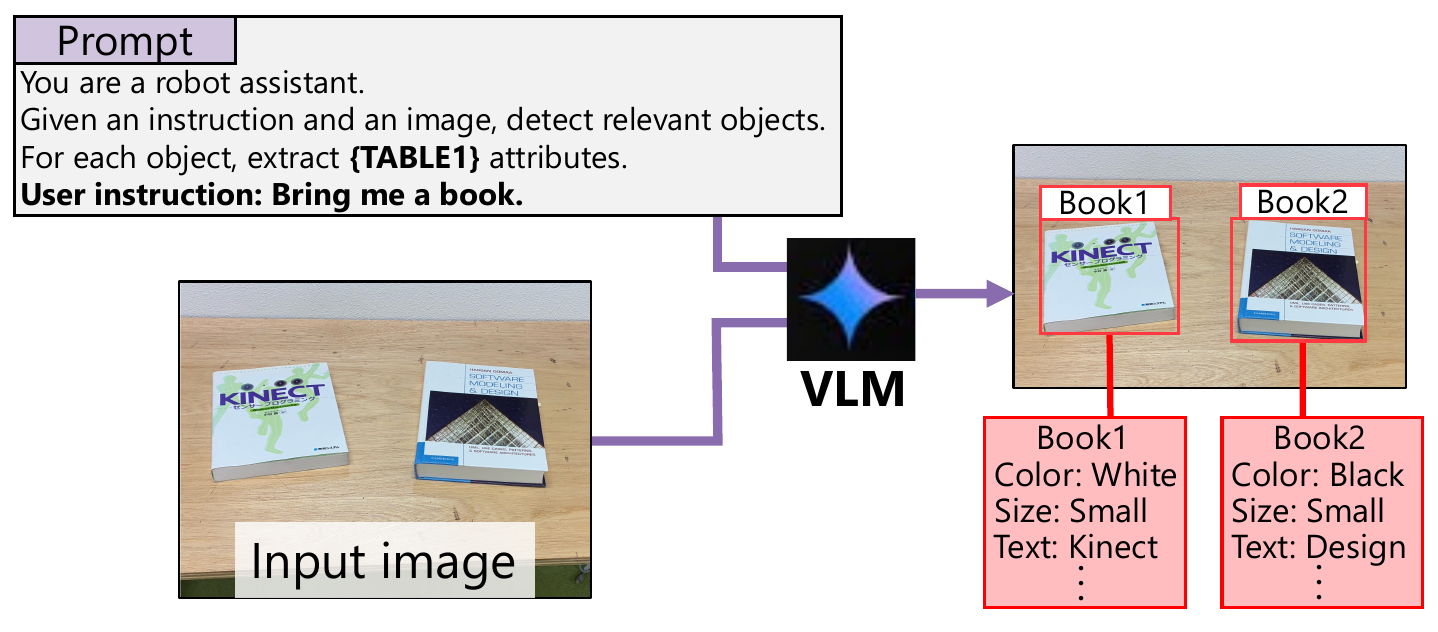} 
   \caption{Overview of target candidate and visual attribute extraction.}
   \label{question_A}
\end{figure}

\section{PROPOSAL}

The proposed mechanism aims to uniquely identify the target object through conversation in response to ambiguous object search instructions, given only the target category.

In this section, we define the visual attributes used in the proposed mechanism, then describe the details of each component.
Visual attributes are features extracted from the appearance of objects that distinguish among candidate objects.
Table~\ref{tab:visual_attributes_definition} shows the visual attributes used in this study.
We selected these attributes based on features considered important in human object identification~\cite{treisman1980feature, salbench}.

Fig.~\ref{proposed_question} shows the overview of the proposed mechanism.
The proposed mechanism extracts candidate objects from the user instruction.
Simultaneously, the proposed mechanism extracts the visual attributes of each candidate (Fig. 2. A).
When the target object is not uniquely determined, the proposed mechanism generates a question to resolve the ambiguity.
Next, the proposed mechanism selects the visual attribute that most effectively identifies the target object from the candidate set (Fig. 2. B).
When multiple visual attributes have the same IG, the proposed mechanism applies a filter based on cognitive characteristics (Fig. 2. C).
The filter determines the visual attribute that users can most intuitively answer.
Finally, the proposed mechanism re-inputs to the VLM a prompt instructing it to generate a concise question about the selected attribute.


\subsection{Extraction of Candidate Objects and Visual Attributes}

Fig.~\ref{question_A} shows an overview of the candidate object and visual attribute extraction process in the proposed mechanism.
As shown in Fig.~\ref{question_A}, the VLM receives a prompt containing the user instruction and input images.
The VLM analyzes the image and detects objects of the instructed category as candidate objects.
Simultaneously, the VLM extracts and structures the visual attributes defined in Table~\ref{tab:visual_attributes_definition} from each candidate object.
Using this approach, the proposed mechanism enables the extraction of candidate objects, such as Book1 and Book2, along with structured visual attribute information for each candidate.
To generate a question, we first need to obtain the visual attributes of each candidate object.
This requires two functions: detecting the objects in the image, and extracting their attributes.
A VLM provides both functions at once.
We therefore adopt a VLM in our mechanism.
The VLM detects the candidate objects and extracts their attributes directly from raw images.
It needs no category-specific detectors or hand-crafted attribute extractors.
It can also render a selected attribute into a natural-language question.
The contribution of this study is therefore not the use of the VLM itself.
It is the mechanism that controls which attribute the VLM is asked about.

The purpose of the proposed mechanism is to design a question generation algorithm based on information gain and cognitive characteristics.
Therefore, objects that the VLM fails to detect are excluded from both the candidates and the ground truth.
In this way, we evaluate the question generation algorithm independently of object-detection accuracy.

\subsection{Visual Attribute Evaluation Based on Information Gain}

$C = \{c_1, c_2, \ldots, c_N\}$ denote the candidate set comprising the candidate objects extracted by the VLM.
To narrow down the target object from
$C$, the proposed mechanism must determine which visual attribute to ask about in order to most efficiently reduce uncertainty.
In this study, we use IG as the evaluation metric for efficiently reducing uncertainty.
IG represents the expected reduction in uncertainty obtained by asking a question about attribute 
$a$.
The VLM extracts candidate objects based on the user's natural language instruction (e.g., ``bring me a book'').

Therefore, this study assumes that the probability $p$ of each candidate object $c_i$ in the candidate set $C$ being the target object is uniform, defined as
\begin{equation}
    p(c_i) = \frac{1}{N}
    \tag{3.1}
\end{equation}
The uncertainty before questioning is defined as entropy 
$H(C)$ as follows.

\begin{equation}
H(C) = - \sum_{i=1}^{N} p(c_i) \log p(c_i)
\tag{3.2}
\end{equation}
When asking about a visual attribute $a$, we assume the user's answer is one of the values of $a$ for each candidate object.
$V_a$ denotes the set of these values (e.g., for Color in Fig.~\ref{question_A}, $V_a = \{$White, Black$\}$)
The following equation expresses the conditional entropy after questioning.

\begin{equation}
H(C \mid a)
=
\sum_{v \in V_a}
p(v) H(C \mid a = v)
\tag{3.3}
\end{equation}
$p(v)$ represents the proportion of candidate objects in the candidate set $C$ that have attribute value $v$,
$H(C \mid a = v)$ is the entropy of the candidate set given that attribute value $v$ is observed.
The $IG(a)$ is defined as the difference in entropy before and after questioning by the following equation.

\begin{equation}
IG(a) = H(C) - H(C \mid a)
\tag{3.4}
\end{equation}

A larger value of $IG(a)$ indicates that a question about attribute $a$ reduces the uncertainty of the candidate objects.

In the example shown in Fig.~\ref{proposed_question}, Color and Text have different values among the candidates, resulting in high IG.
On the other hand, when Size is common across all candidates, IG of Size becomes low.
The proposed mechanism computes IG for all predefined visual attributes and selects the attribute with the highest discrimination efficiency.
When the target object cannot be uniquely identified with a single question, the proposed mechanism updates the candidate set based on the user's answer and recomputes IG.
By repeating this process, the proposed mechanism selects the optimal attribute at each dialogue turn.

\subsection{Visual Attribute Filter Based on Cognitive Characteristics}

IG is a metric that quantitatively evaluates the ease of distinguishing among candidate objects.
However, IG measures only the reduction in uncertainty and does not consider the ease of response or intuitive recognition for users.
Even when an attribute has high discrimination efficiency, asking about features that users cannot recall may cause hesitation in responses and prevent smooth dialogue.

Therefore, we introduce a ``visual attribute filter'' based on a preliminary user questionnaire.
The filter is a ranked list of visual attributes that humans prioritize when identifying objects, defined for each object category.
The details of the questionnaire are described in Section~\ref{sec:ex_qu}.

\subsection{Question Generation Algorithm}
In this section, we explain the question generation algorithm that integrates discrimination efficiency based on IG and cognitive ease based on the visual attribute filter.

First, the proposed mechanism computes IG of each visual attribute for the candidate object set.
In this step, attributes that cannot partition the candidate set, attributes where
$IG(a)=0$, are excluded.
Next, the proposed mechanism selects the attribute with the highest IG.
When multiple attributes have the same IG, the proposed mechanism refers to the visual attribute filter and selects the attribute with the highest cognitive priority.
Then, the proposed mechanism provides the selected attribute name and the attribute value list for each candidate object as prompts to the VLM.
The VLM generates a concise natural-language question about the selected single attribute.
For example, when Color is selected, a question such as ``Which book would you like, the red, orange, or white one?'' is generated.

By following the steps below, the proposed mechanism achieves both discrimination efficiency and ease of user response, enabling efficient target identification for ambiguous user instructions.

\section{EXPERIMENTAL SETTINGS}

This section describes the experimental setup for evaluating the effectiveness of the proposed adaptive question generation mechanism.
We demonstrate the effectiveness of the proposed mechanism from two perspectives, assuming ambiguous natural language instructions: (1) the efficiency of ambiguity resolution through dialogue, and (2) the ease of response to questions for users.
We conducted large-scale user experiments targeting a broad range of general users through crowd-sourcing.

\subsection{Dataset Construction}
\label{dataset_construction}

\begin{figure}[tb]
 \centering
 \includegraphics[width=\linewidth]{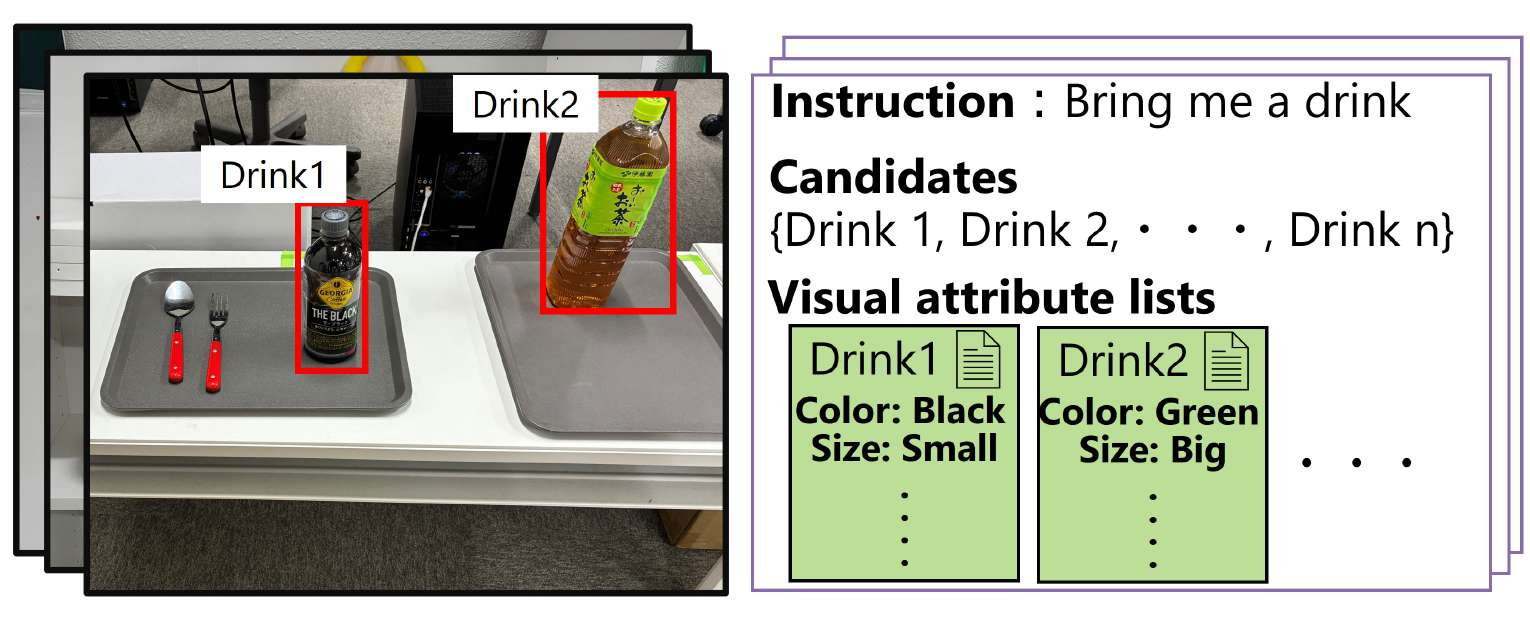}
 \caption{Dataset examples: Scenes containing ambiguous objects in a university campus environment.}
 \label{dataset}
\end{figure}


We collected images in indoor environments on a university campus and constructed an evaluation dataset for this evaluation.
The dataset consists of 120 scenes where multiple everyday objects are arranged.
The dataset organizes each scene as a set consisting of an ambiguous instruction and a corresponding candidate object set.
Fig.~\ref{dataset} shows an example of the scene structure.
Each scene consists of an input image, a user instruction, a candidate object set of the instructed category, and the visual attributes of each candidate.
As shown in Fig.~\ref{dataset}, for an instruction such as ``bring me a drink,'' the dataset defines a situation in which multiple drink candidates are available.
$C=\{Drink_1,Drink_2,\dots,Drink_n\}$.
Each candidate object is annotated with six visual attributes, in addition to bounding-box information: Color, Size, Shape, Text, Material, and Illustration.
In constructing the dataset, we used Gemini 3.0 Pro~\cite{google_gemini3pro_modelcard} throughout, from extracting bounding box coordinates of candidate objects to extracting visual attributes.
Subsequently, we conducted manual verification to ensure scene reliability and defined the ground truth by excluding false detections and inaccurate attribute descriptions.

The purpose of the proposed mechanism is to design a question generation algorithm based on information gain and cognitive characteristics.
Therefore, objects that the VLM fails to detect are excluded from both the candidates and the ground truth.
In this way, we evaluate the question generation algorithm independently of object-detection accuracy.

\begin{figure}[tb]\centering\includegraphics[width=\linewidth]{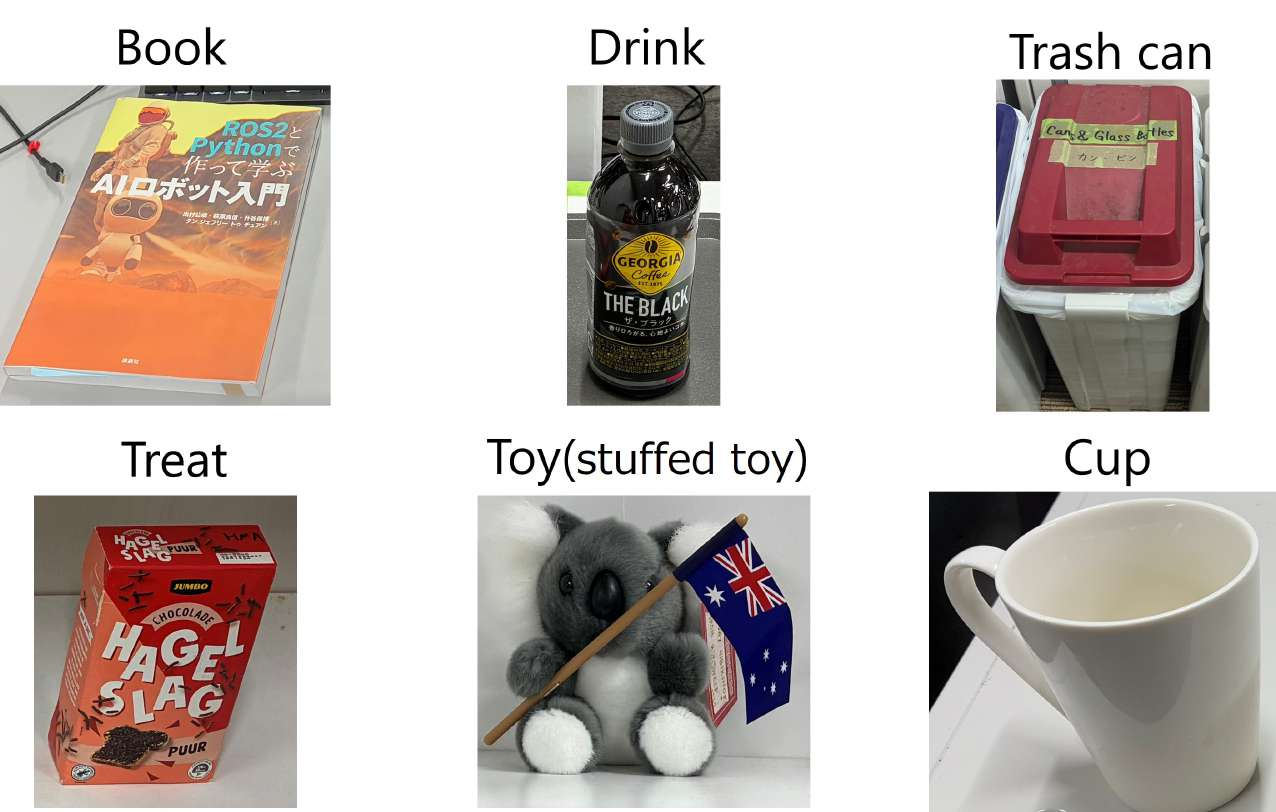}
\caption{Examples of objects in each of the six categories.}
\label{ex_obj}
\end{figure}

\subsection{Determination of Visual Attribute Filter}
\label{sec:ex_qu}

\begin{figure}[tb]\centering\includegraphics[width=\linewidth]{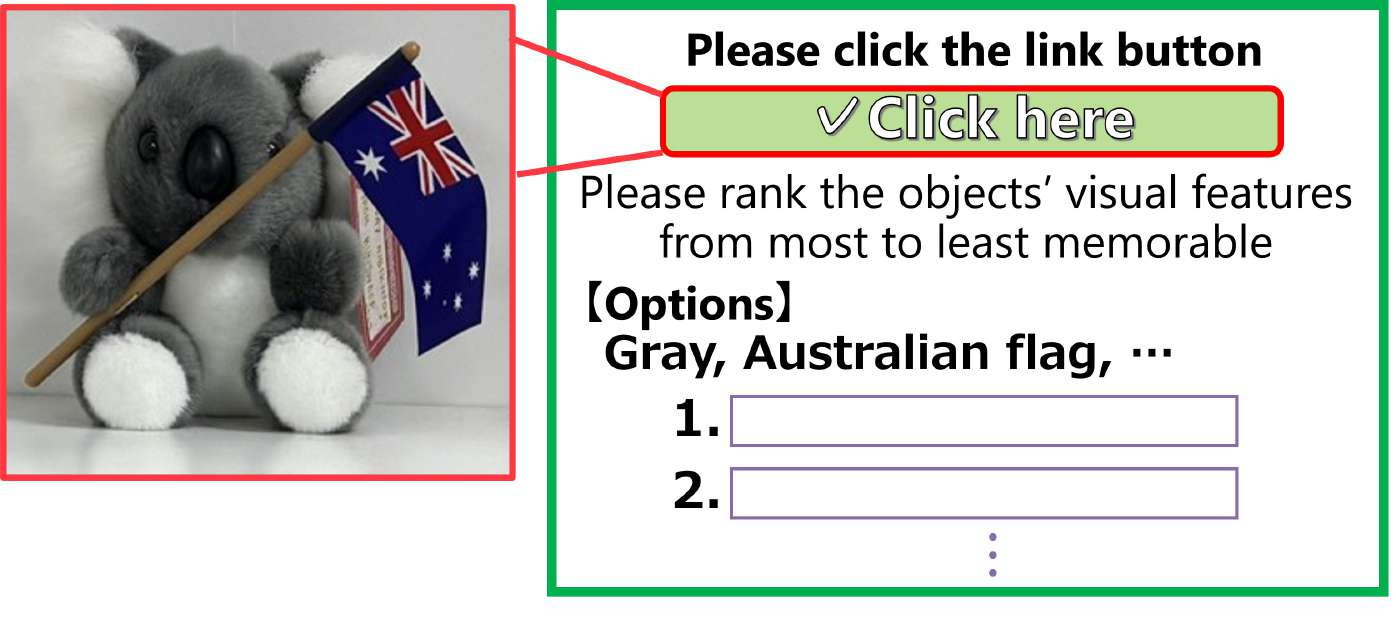}
\caption{Example of the web-based questionnaire for visual attribute ranking. Green area: the questionnaire interface, where participants rank the memorability of visual features. Red area: clicking the link button displays the target object image.}
\label{web}
\end{figure}

In the proposed mechanism, we need to quantify cognitive load and response ease for each visual attribute.
To determine the visual attribute filter used in this evaluation, we conducted a large-scale web questionnaire survey targeting a broad range of general users.
We targeted six object categories that frequently appear in the dataset (book, drink, trash can, treat, toy, cup) and collected responses from 1,200 participants.
Respondents on the crowdsourcing platform participate largely anonymously.
We could not reliably collect detailed demographics such as age and gender. 
Fig.~\ref{ex_obj} shows examples of each object category.
Fig.~\ref{web} shows an example of the questionnaire.
Participants answered the questionnaire according to the following steps.

\begin{itemize}
\item \textbf{1. Observation:} Participants press a button to display the target image to view an image of the target object. They are instructed to intuitively memorize the visual features of the object.
They then close the image to ensure that subsequent ranking is based solely on memory.
\item \textbf{2. Ranking:} Participants receive the attribute values of the target object extracted in Section~\ref{dataset_construction} (e.g., Gray as a Color value, Australian flag as an Illustration value defined in Table I, ...) as options.
Participants ranked the attributes in order of how memorable and useful they found them for identification.
\end{itemize}

\subsection{Evaluation Metrics}
In this subsection, we explain the evaluation metrics used to assess the effectiveness of the proposed question-generation mechanism across large-scale crowdsourcing and a fixed dataset.
We conducted the evaluation targeting a broad range of general users via Yahoo! Crowdsourcing.
Real-time question generation is difficult in a crowdsourcing environment.
Therefore, we focus on the quality of the first question generated by the proposed mechanism.
This evaluation is therefore single-turn by design.
Our aim is to isolate the effect of the cognitive attribute filter on a single generated question. Specifically, we measure whether it improves discrimination efficiency and ease of response. 

We quantitatively evaluate two aspects: discrimination efficiency, indicating whether the target can be identified in a single question-answer turn, and ease of response for users.
We selected samples from the dataset in which the target can be uniquely identified when an appropriate attribute is chosen.
The evaluation metrics are as follows:

\begin{enumerate}
\item \textbf{Identification Success Rate:} The proportion of cases in which the VLM correctly identified the target object based on the user's answer.
We input the dialogue history into the VLM to estimate the target region and considered a case successful when the Intersection over Union (IoU) with the ground truth region was 0.8 or above.
\item \textbf{Mean Rank Score:} The average ranking assigned by participants. 
A lower value indicates a more intuitive and easier-to-answer question.
\end{enumerate}

In the dataset, multiple candidate objects are located in close proximity within the same image.
Therefore, we adopted an IoU threshold of 0.8, which is stricter than the commonly used object detection threshold of 0.5, to prevent incorrectly accepting adjacent candidates as the ground truth.

\subsection{Comparison Methods}

We evaluate the effectiveness of the questions generated by the proposed mechanism using the constructed dataset of 120 scenes.
As discussed in the related works, question strategies based on probabilistic discrimination efficiency have been shown to be effective for ambiguity resolution.
However, whether incorporating human cognitive characteristics into the question strategy further improves performance remains to be seen.
To verify the effectiveness of incorporating human cognitive characteristics into question strategy, we set the following three comparison methods.

\textbf{Standard VLM:} A method that allows the VLM to freely generate questions without any constraints on visual attributes.
By comparing with this method, we verify whether explicit attribute control based on both discrimination efficiency and cognitive characteristics is necessary over unconstrained question generation.

\textbf{Attribute-Selected VLM:} A method that constrains the VLM to select a single visual attribute for questioning, but delegates the attribute selection itself to the VLM.
By comparing with this method, we verify whether the VLM can select attributes that are intuitively easy for users to answer when multiple attributes have equivalent discrimination efficiency.

\begin{table}[tb]
\centering
\caption{Average Rank of Visual Attributes by Object Category (Lower score indicates higher cognitive priority).}
\label{tab:attribute_rank_mean}
\begin{tabular}{lcccccc}
\toprule
Attribute & Book & Drink & Trash & Treat & Toy & Cup \\
\midrule
Text         & \textbf{1.59} & \textbf{1.65} & 2.14 & \textbf{2.12} & -- & 2.60 \\
Color        & 2.69 & 2.48 & \textbf{1.98} & 2.59 & 2.25 & \textbf{1.45} \\
Illustration & 2.80 & 3.36 & 2.58 & 3.01 & 2.54 & 2.03 \\
Shape        & 4.10 & 3.66 & 3.56 & 3.68 & \textbf{2.06} & 2.05 \\
Material     & 4.67 & 4.01 & 4.12 & 4.41 & 2.95 & 3.19 \\
Size         & 5.60 & 5.46 & 5.71 & 5.80 & 4.71 & 4.51 \\
\bottomrule
\multicolumn{7}{l}{\footnotesize *Bold values indicate the highest priority (lowest rank).}\\
\multicolumn{7}{l}{\footnotesize *``Toy'' category did not contain text attributes.}
\end{tabular}
\end{table}
\textbf{Proposed:} 
The proposed mechanism selects the optimal attribute and generates questions by integrating discrimination efficiency based on IG and ease of response based on cognitive characteristics obtained from a preliminary experiment.

We used Gemini 3.0 Pro as the VLM for all methods.
In each trial, we presented participants with an image and an ambiguous instruction, along with the questions generated by the three methods described above.
Participants answered the three presented questions and ranked them from first to third place in terms of ease of response.




\section{EXPERIMENTAL RESULTS}


\subsection{Visual Attribute Filter}
Attributes that users can intuitively recall are considered to enable prompt responses when asked about them.
Therefore, we use the questionnaire's ranking as an indicator of ease of response.
Table~\ref{tab:attribute_rank_mean} shows the mean rank of each visual attribute computed from the questionnaire responses.
Overall, Text and Color ranked highly across many categories.
This ordering is consistent with prior accounts of saliency. Perceptually salient features such as color and text are recalled preferentially~\cite{treisman1980feature, salbench}. 
This supports interpreting the survey ranking as empirically grounded rather than arbitrary.
In particular, Text was prioritized for packaged products, while Color was prioritized for objects with little textual information.
As an exception, Shape was prioritized most highly in the toy category, which contains no textual information.
On the other hand, Material and Size consistently ranked low across all categories.
In this experiment, the Toy category targets stuffed animals and consists of objects with no textual information.

\begin{table*}[tb]
\centering
\caption{Examples of Questions Generated by Each Method for Book, Toy, and Drink Categories.}
\label{tab:question_examples}
\footnotesize
\setlength{\tabcolsep}{4pt}
\renewcommand{\arraystretch}{1.3}
\begin{tabularx}{\linewidth}{l >{\raggedright\arraybackslash}X >{\raggedright\arraybackslash}X >{\raggedright\arraybackslash}X}
\toprule
Category & \textbf{Standard VLM} & \textbf{Attribute-Selected VLM} & \textbf{Proposed Method} \\
\midrule
\textbf{Book}
& There are several books on the shelf. I can also see an OpenCV book. Which one should I bring?
& Which color book would you like?
& Which book would you like: the red, orange, or white one? \\
\midrule
\textbf{Toy}
& I see several stuffed animals. There is a koala, a white character, a hedgehog, and a dog. Which one should I bring?
& Which one is better: the koala, the white character, the hedgehog, or the brown dog?
& Which stuffed animal would you like: the gray, white, or brown one? \\
\midrule
\textbf{Drink}
& I can see several drinks: Aquarius, Red Bull, black coffee, and green tea. Which one should I bring?
& Which type of drink would you like: a sports drink, an energy drink, coffee, or tea?
& Which would you like: Aquarius, Red Bull, Georgia The Black, or green tea? \\
\bottomrule
\end{tabularx}
\end{table*}

\begin{figure*}[tb]\centering\includegraphics[width=\linewidth]{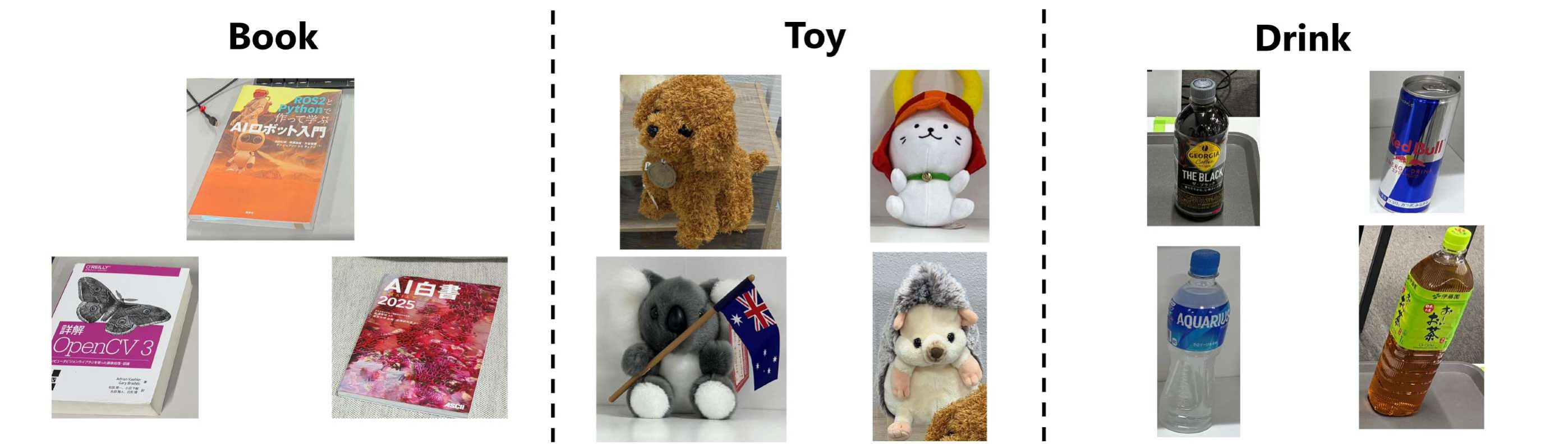}
\caption{Examples of candidate objects for the three categories shown in Table~\ref{tab:question_examples}. 
All objects in this figure are treated as possible candidates during the questioning process, from which questions are generated to identify the target object.}
\label{candidates}
\end{figure*}

\subsection{Question Generation}
Table~\ref{tab:question_examples} shows examples of questions generated by three methods.
Fig.~\ref{candidates} shows the candidate objects for the three categories used in Table~\ref{tab:question_examples}, from which questions are generated to identify the target object.
As shown in Table~\ref{tab:question_examples}, Standard VLM tends to produce longer questions.
These questions combine several attributes and object names. Combining attributes can aid clarity in some contexts.
In our setting, a single high-information-gain attribute already resolves the ambiguity.
This attribute is also easily recalled. 
Attribute-Selected VLM focuses on a single attribute, but does not always select an appropriate attribute.
The proposed mechanism generates concise questions focused on attributes with high discrimination efficiency that users can easily answer.

Table~\ref{tab:exp1_results} shows the target identification success rate and the subjective evaluation results of ease of response.
The proposed mechanism achieved the highest target identification success rate of 88.5\%.
The 79.4\% success rate for Standard VLM indicates that the proposed approach contributed to reliable target identification.
On the other hand, the success rate of Attribute-Selected VLM remained at 73.7\%.
This result suggests that selecting attributes without considering discrimination efficiency among candidate objects risks generating questions that are difficult to understand.
Notably, Attribute-Selected VLM (73.7\%) performs below Standard VLM (79.4\%), despite both selecting questions with a VLM.
Regarding ease of response, the proposed mechanism recorded a mean rank of 1.36, receiving strong support compared to the other methods.

These results demonstrate that selecting a single visual attribute based on probabilistic discrimination efficiency enables efficient target identification and that incorporating human cognitive characteristics generates questions that are easy for users to answer.
These findings quantitatively support the hypothesis that ``integrating both factors enables the generation of questions that achieve high discrimination efficiency and are easy for users to answer.''

\begin{table}[tb]
\centering
\caption{Quantitative Results of Experiment 1: Identification Success Rate and Subjective Mean Rank Score}
\label{tab:exp1_results}
\begin{tabular}{lcc}
\toprule
Method & Success Rate (\%) $\uparrow$ & Mean Rank Score $\downarrow$ \\
\midrule
Standard VLM            & 79.4 & 2.67 \\
Attribute-Selected VLM  & 73.7 & 1.97 \\
\textbf{Proposed}       & \textbf{88.5} & \textbf{1.36} \\
\bottomrule
\multicolumn{3}{r}{\footnotesize *Mean Rank Score: Lower is better (1: Best, 3: Worst).}
\end{tabular}
\end{table}

\section{DISCUSSION}
The proposed mechanism achieved the highest target identification rate and the best subjective evaluation by integrating mathematical discrimination capability and ergonomic cognitive ease.
As shown in Table~\ref{tab:question_examples}, Standard VLM generates redundant questions containing multiple attributes.
For example, in the Book category, Standard VLM generates complex questions that refer to specific book titles.
The increase in users' cognitive load is considered the cause of the low evaluation.
On the other hand, Attribute-Selected VLM selects a single attribute, color, in the Book category and generates a concise question.
However, in the Toy category,  Attribute-Selected VLM generates questions that enumerate object names rather than visual attributes.
This suggests that the failure to appropriately select attributes effective for identification led to the lower success rate.
More specifically, this drop is caused not by the absence of information gain, but by the lack of a predefined set of attributes. 
In Attribute-Selected VLM, the VLM defines the attributes itself, so it may select attributes that are unsuitable for distinguishing the candidates.
The proposed method instead defines the attributes in advance, based on prior work, and known to be easy for humans to recognize.
The proposed mechanism selects attributes from candidates based on IG, then selects the attribute with the highest cognitive priority.
In the Toy category, the proposed mechanism selects color as an attribute with high discrimination efficiency, enabling users to answer and generate a concise question easily.
These results indicate that considering not only probabilistic discrimination efficiency but also human cognitive characteristics is important in question design.

While these results support our hypothesis, our evaluation is a controlled study.
The main direction of future work is therefore deployment on a real robot.
Such deployment must handle detection errors, since we assumed perfect detection.
It must also replace crowdsourcing and pre-viewed images with live interaction and memory-based recall.
In addition, it must extend our single-question evaluation to multi-turn dialogue, including recovery from incorrect answers.
Combining the mechanism with an ambiguity-detection approach such as KnowNo is a promising step.
A further limitation is scope: our participants were Japan-centric and from a single campus.
The generalizability of the attribute rankings is therefore limited.
However, the attributes we rely on, such as color, shape, and text, are perceptually salient features.
Such salience is regarded as broadly perceptual rather than culture-specific~\cite{treisman1980feature, salbench}.
Thus, the advantage of the proposed method is expected to hold beyond this population.

\section{CONCLUSIONS}
In this study, we proposed an adaptive question-generation mechanism that uniquely identifies the target object via dialogue in response to ambiguous object-search instructions.
The proposed mechanism integrates visual attribute evaluation based on information gain (IG) and an attribute filter based on human cognitive characteristics.
By integrating both factors, the proposed mechanism enables the generation of questions with high discrimination efficiency that users can easily answer.
As a result, the proposed mechanism achieved a high target identification success rate and received high ratings in the subjective evaluation of ease of response from the user.
These results demonstrate the importance of considering not only probabilistic discrimination efficiency but also human cognitive characteristics in the design of questions.
The experiments in this study supported the hypothesis that ``integrating probabilistic discrimination efficiency and human cognitive characteristics enables the generation of questions that achieve high discrimination efficiency and are easy for users to answer.''

In future work, combining the proposed mechanism with ambiguity detection approaches such as KnowNo is a promising direction.
In particular, evaluating the proposed mechanism in a multi-turn setting, measuring the convergence rate over successive questions, and the robustness to incorrect user answers, is the primary direction for future work.
This combination is expected to yield an ambiguity-resolution system that considers both the timing and the content of questions.
As a future challenge, the visual attribute filter used in this study targets six object categories; extending it to more diverse object categories is necessary.
Furthermore, the dataset used in this study was collected in indoor environments on a university campus.
Therefore, verification of generalization performance across different environments, such as homes and commercial facilities, is required.
Finally, we aim to realize question generation adapted to individual users by collecting large-scale dialogue data and modeling cognitive characteristics in a general manner using preference learning.

\addtolength{\textheight}{-12cm}   





\bibliographystyle{ieeetr}
\bibliography{ref.bib}

\end{document}